\pdfoutput=1

\documentclass[11pt]{article}

\usepackage[]{emnlp2022}

\usepackage{times}
\usepackage{latexsym}

\usepackage{hyperref}
\usepackage{graphicx}
\usepackage{amsmath}
\usepackage{multirow}
\usepackage{array}
\usepackage{algorithm}
\usepackage{algpseudocode}
\usepackage{pifont}
\newcommand{\cmark}{\ding{51}}%
\newcommand{\xmark}{\ding{55}}%

\usepackage[T1]{fontenc}

\usepackage[utf8]{inputenc}

\usepackage{microtype}

\title{Entity Aware Negative Sampling with Auxiliary Loss of False Negative Prediction for Knowledge Graph Embedding}

\author{Sang-Hyun Je \\
  Kakao Enterprise Corp. \\
  \texttt{shje65@gmail.com} \\}

\begin{document}
\maketitle

\begin{abstract}
Knowledge graph (KG) embedding is widely used in many downstream applications using KGs.
Generally, since KGs contain only ground truth triples, it is necessary to construct arbitrary negative samples for representation learning of KGs. Recently, various methods for sampling high-quality negatives have been studied because the quality of negative triples has great effect on KG embedding. In this paper, we propose a novel method called Entity Aware Negative Sampling (EANS), which is able to sample negative entities resemble to positive one by adopting Gaussian distribution to the aligned entity index space. 
Additionally, we introduce auxiliary loss for false negative prediction that can alleviate the impact of the sampled false negative triples. The proposed method can generate high-quality negative samples regardless of negative sample size and effectively mitigate the influence of false negative samples. The experimental results on standard benchmarks show that our EANS outperforms existing the state-of-the-art methods of negative sampling on several knowledge graph embedding models. Moreover, the proposed method achieves competitive performance even when the number of negative samples is limited to only one.
\end{abstract}

\section{Introduction}

Knowledge graph (KG) is a multi-relational directed graph that contains various entities and their relationships.
\begin{figure}[h]
    \centering
    \includegraphics[width=\columnwidth]{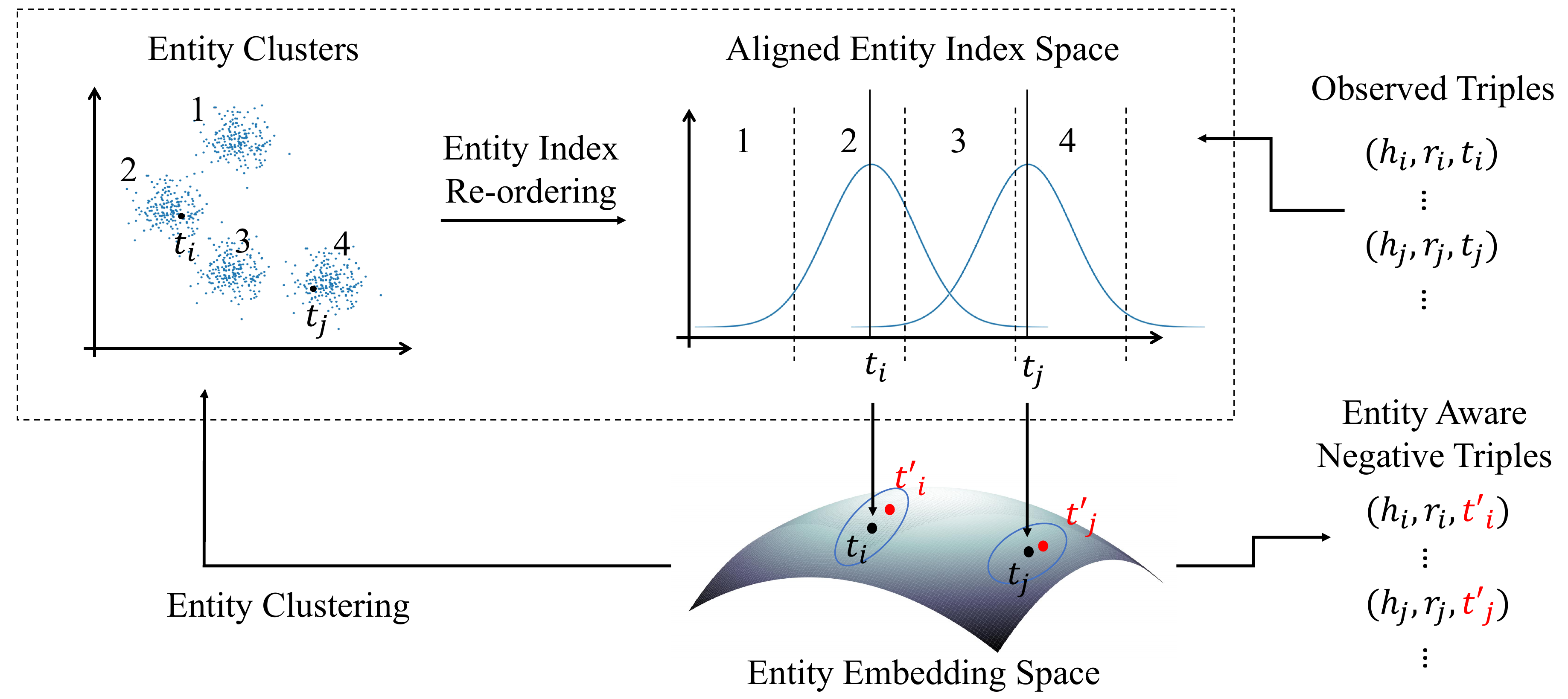}
    \caption{The overview of proposed entity aware negative sampling method.}
    \label{fig:eans}
\end{figure}
Each edge of KG describes factual information, called a \texttt{fact} or \texttt{triple}. A fact as triple is composed of two entities and relation corresponding to their relationship and it is represented in the form of \texttt{(head entity, relation, tail entity)}, which is denoted as \texttt{(h, r, t)}, e.g. \texttt{(Christopher Nolan, DirectorOf, Interstellar)}. Freebase \cite{bollacker2008freebase}, YAGO \cite{suchanek2007yago}, and DBpedia \cite{auer2007dbpedia} are examples of large scale knowledge graphs that contains real world information. Recently, KG is being actively used to inject structured knowledge into target system in various fields such as recommendation \cite{wang2019kgat, xu2020product, zhang2018learning}, question answering \cite{huang2019knowledge, saxena2020improving}, and natural language generation \cite{liu2021kg, wu2020diverse}. 

KGs are usually incomplete because there are inevitably missing links between entities. 
To predict these missing link in KGs, a lot of link prediction (a.k.a., knowledge graph embedding) models are trying to embed elements of KG to low dimensional vector space. Since KGs basically contain only true triples, most of the existing knowledge graph embedding model are trained in a contrastive learning manner that widens the gap of scores between true and false triples.

Obviously, the quality of negative samples is critical to learning knowledge graph embedding. 
Nevertheless, replacing head or tail entity of positive triple with random entity from the entire entities of KG for constructing negative samples is the most widely used because of its efficiency. Such a uniform random sampling method can be effective at the beginning of training. But as training progresses, the trivial negative samples lose their effectiveness and give zero loss to the training model  \cite{wang2018incorporating}. To resolve this problem, many studies have been proposed to construct more meaningful negatives.

To generate high-quality negative samples, we propose an entity aware negative sampling (EANS) method that exploits entity embeddings of the knowledge graph embedding model itself. EANS can generate high-quality negatives regardless of the negative sample size. The proposed method constructs negative samples based on the assumption that \textit{the negative triples which are corrupted by similar candidate entities to the original positive entity will be high-quality negative samples}. EANS samples negative entities similar to positive one by utilizing the distribution of entity embeddings. The generated entity aware negative samples push the model to continuously learn effective representations.

While generating high-quality negative samples, it should be careful of the influence of false negatives. 
When corrupting positive triples using entities which are similar to the positive one, the possibility of generating false negative samples also increases. To alleviate the effect of false negative triples, we propose an auxiliary loss for false negative prediction. The proposed function can mitigate the effect of false negatives by calculating additional prediction scores and reducing the triple scores of false negatives.

We evaluate the proposed EANS method on several famous knowledge graph embedding models and two widely used benchmarks. Based on the experimental results, proposed method achieves remarkable improvements over baseline models. Moreover, it shows better performance than the existing negative sampling methods on several models. Our method also shows comparable performance while using much smaller number of negative samples. Especially EANS produces competitive performance to existing negative sampling methods even with only one negative sample.

\section{Related Work}

\subsection{Knowledge Graph Embedding Models}
There are two main streams in the knowledge graph embedding, one for translational distance models and the other for semantic matching models. TransE \cite{bordes2013translating} is the first proposed translational distance based model. Various extensions of TransE, such as TransH \cite{wang2014knowledge}, TransR \cite{lin2015learning} and TransD \cite{ji2015knowledge}, increase their expressive power by projecting entity and relation vectors into various spaces. RESCAL \cite{nickel2011three}, DistMult \cite{yang2014embedding}, and ComplEx \cite{trouillon2016complex} is the most representative models of semantic matching based methods. RESCAL treats each relation as a matrix which capture latent semantics of entities. DistMult simplifies RESCAL by constraining relation matrices to diagonal matrices. ComplEx is an extension of DistMult that extend embedding vectors into complex space. Recently, more complex and sophisticated models \cite{dettmers2018convolutional, vashishth2019composition, sun2019rotate, lu2022dense, vashishth2020interacte} have been studied. Such methods introduce various technique and networks to model the scoring function and extend embeddings of KG's elements into various spaces.

\subsection{Negative Sampling}
To construct meaningful negative samples, \cite{wang2018incorporating, cai2017kbgan} proposed Generative Adversarial Network(GAN) \cite{goodfellow2014generative} based architecture to model the distribution of negative samples. However, these methods need much more additional parameters for extra generator and it could be hard to train GAN because of its instability and degeneracy \cite{zhang2019nscaching}. To address these problems, caching based method \cite{zhang2019nscaching} have been proposed with fewer parameters compared to GAN-based methods to keep high-quality negative triples. \cite{ahrabian2020structure} suggested the method that utilize the structure of graph by choose negative samples from k-hop neighborhood in the graph. \cite{sun2019rotate} proposed self-adversarial negative sampling, which give difference weights to each sampled negative according to its triple score. In recent, \cite{hajimoradlou2022stay} proposed different training procedure without negative sampling. Instead of using negative samples, they fully utilize regularization method to train knowledge graph embeddings.
\section{Method}

In this section, we introduce our proposed entity aware negative sampling (EANS) method.
The proposed method consists of two parts. 
\begin{algorithm}
\textbf{Input}: Knowledge graph $\mathcal{G} = \{(h, r, t)\}$, entity set $\mathcal{E}$, relation set $\mathcal{R}$
\begin{algorithmic}[1]
    \State Initialize embeddings $W^{e}$ for each $e \in \mathcal{E}$ and $W^{r}$ for each $r \in \mathcal{R}$
    \For{$i = 1, \ldots, max\_step$}
        \State sample a mini-batch $\mathcal{G}_{batch} \in \mathcal{G}$
        \For{$(h, r, t)\in \mathcal{G}_{batch}$} 
            \State get negative entity $h'$(or $t'$), where \newline 
            \hspace*{3em} $h' = int(h$(or $t$) $+\mathcal{N}(0, 1) * \sigma)$
            \State construct negative triple $(h', r, t')$
            \State update parameters w.r.t. the gradients \newline 
            \hspace*{3em} of loss function, Eq. \ref{eqn:final_loss}.
        \EndFor
        \If{($i$ mod $reorder\_step$) == $0$}
            \State clustering the entity embeddings $W^{e}$ \newline
            \hspace*{3em} using K-means method
            \State re-ordering the index of $\mathcal{E}$ based on \newline
            \hspace*{3em} clustering labels
        \EndIf
    \EndFor
\end{algorithmic}
\caption{Algorithm of EANS}
\label{alg:eans_algorithm}
\end{algorithm}
One is selecting a negative entity by re-ordering entire entities to create entity aware negative triples, and the other is calculating an additional loss to mitigate the influence of false negatives. The remainder of this section gives details of each part.
Entire process of proposed method are summarized in Algorithm \ref{alg:eans_algorithm}.

\subsection{Entity Aware Negative Sampling (EANS)}

\subsubsection{Entity Embedding based Clustering}

Given a KG, let $ \mathcal{E} $ be the entire entities set, $ \mathcal{R} $ be the relation set, and $ \mathcal{G} $ be all truth triple sets. A triple score $f(h, r, t)$ is calculated by an adopted specific knowledge graph embedding model.

In the general uniform random negative sampling method, a negative triple, $ (h', r, t)$ or $ (h, r, t') $, can be constructed by corrupting the entities of an observed positive triple $ (h, r, t) $, where $h', t' \in \mathcal{E}$. When corrupting the entities, random entities are extracted from uniformly weighted $ \mathcal{E} $. Since the most of the entities in $ \mathcal{E} $ are not highly related to the each positive triple $(h, r, t)$, it is hard to expect sampling high-quality entities through the uniform random sampling method. 

In order to sample meaningful entities, we design EANS to select negative entities that are highly related to the positive one. The key intuition of our method is that \textit{two entities which have similar embedding vectors can be high-quality negative sample to each other}. Therefore, it is possible to construct high-quality negative samples by corrupting with entities that have similar embedding vectors to positive entity.

The simplest way to find entities similar to the positive entity is calculating the distances between the embedding vectors of the positive entity and entire entities in $\mathcal{E}$ at every training step and searching the nearest neighbors. However, this method consumes a large computational resource and the cost will increases dramatically in proportion to the size of embedding dimension $d$ and entities set $\mathcal{E}$.

Instead of applying the nearest neighbor searching for negative entity selection at every step, we design entity clustering based sampling method. First, our method groups similar entities in advance and samples negative entities based on the clusters. This method can be implemented through the K-means \cite{lloyd1982least} clustering algorithm. Each entity representation $e_{i}$ of $i$-th entity for K-means clustering is constructed by concatenating all entity specific parameters,
\begin{equation}
\label{eqn:ent_repr}
    e_{i} = [W_{i}^{1};W_{i}^{2};\cdots;W_{i}^{L}],
\end{equation}
where $W^{l}$ are entity specific parameters in model and $;$ denotes concatenating operation.
For example, entity representation of TransD \cite{ji2015knowledge} consists of entity specific embedding and entity transfer vector, and can be represented as $e_{i} = [W_{i}^{emb};W_{i}^{transfer}]$. Given a positive entity, a negative entity is chosen based on the cluster to which the positive entity belongs. The important point of selecting a negative entity is that it is not only selected from the same cluster positive entity belong to, but also selected from the outside of the corresponding cluster. The more details of the method are described in section \ref{sec:method_gaussian}.

Since clustering algorithm is adopted to the entities which are on training for knowledge graph embedding, the entity embeddings will continuously change as learning goes on. Therefore, cluster labels of entities should be constantly updated. However, executing clustering algorithm in every training step is also intractable. Fortunately, clustering algorithm does not need to be executed every training step for negative sampling. In our experiments, it shows sufficient performance even if the cluster labels were only updated every one to three epoch.

\begin{figure}[h]
    \centering
    \includegraphics[width=\columnwidth]{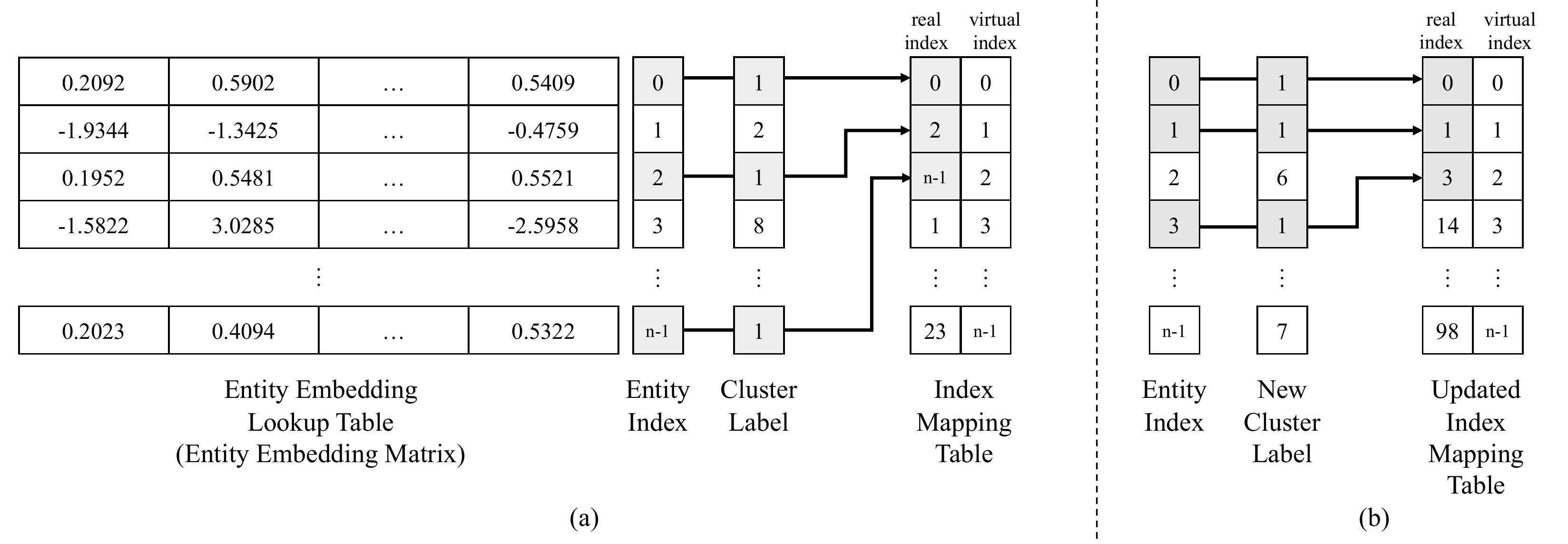}
    \caption{Example of entity index re-ordering. (a) Entities in same cluster can be located far away in the entity matrix, but the virtual indices locate them adjacent to each other. (b) When cluster labels are changed, we only update the virtual-to-real index mapping table. }
    \label{fig:virtual_index}
\end{figure}

\subsubsection{Entity Index Re-Ordering for Gaussian Sampling in Entity Index Space}
\label{sec:method_gaussian}

The K-means clustering algorithm does not guarantee that the numbers of data points in each cluster are evenly divided. If the size of the cluster becomes extremely small, the same entity is used too repeatedly as a negative, which adversely affects learning of models.

To avoid this situation, our method choose negative samples by using index of entity embedding. Commonly, when implementing the negative sampling, one random integer value is uniformly sampled from $[0, size\_of\_entities)$. After then, the entity embedding with this value as an index is fetched. We modify the process by replacing uniform distribution with Gaussian distribution. 

We implement our sampling method to sample a random entity index $x'$ from a Gaussian distribution with variance is $\sigma$ and mean is the index of the positive entity $x$ as $\mu$, .i.e. $ x' \sim \mathcal{N}(\mu, \sigma)$. If the indices of entities with similar embeddings are arranged to be closed, we can draw entity aware random samples from the process. It can be calculated by multiplying the standard normal distribution $\mathcal{N}(0, 1)$ by $\sigma$ and adding $x$ as follows, 
\begin{equation}
\label{eqn:gaussian_distribution}
    x' \sim (\mathcal{N}(0, 1) * \sigma + x),
\end{equation}
where $\sigma$ is hyperparameters that controls the variance of distribution. For given positive triple $(h_{i}, r, t_{j})$, negative samples constructed with corrupted entity, $(h_{i'}, r, t_{j})$ or $(h_{i}, r, t_{j'})$. We can control the quality of negative samples through $\sigma$. If with a smaller value of $\sigma$, more hard negatives will be sampled. On the other hand, if one use a higher value of $\sigma$, more diverse negative entities will be extracted. Figure \ref{fig:eans} shows the overview of EANS.

To get arranged entity indices, entity embeddings have to be re-ordered based on the clustering result. When entity clustering has done, an arbitrary cluster is randomly selected among $K$ clusters, and the index from $0$ to $(N-1)$ is assigned to the entities in this cluster, where $N$ is the size of the cluster. After then, by calculating the distances between the centroid of the current selected cluster and the remaining clusters, take the nearest cluster as the next cluster to assign indices. All entities of the newly selected cluster be assigned to from $N$ to $(2N-1)$-th indices. Repeat this process until all indices of entities are assigned. This re-ordering operation also repeatedly performed according to the changing entity clusters.

For convenience in implementation, we introduce virtual entity indices in the process of entity re-ordering and negative sampling. When the entity cluster labels are updated, only the order of virtual indices and index mapping table are updated. The index mapping table that map the virtual indices and the real entity indices. When negative sampling, a negative entity is selected by using virtual entity indices. All forward operations of model are performed by using the parameters with the real indices connected to the virtual indices of the negatives. An example of entity index re-ordering with virtual index is depicted in Figure \ref{fig:virtual_index}.

\subsection{False Negative Prediction Loss}

Since the EANS generates negative triples using entities similar to a given positive entity, the probability of generating false negative samples can also be increased. To remedy the false negative problem, we introduce a novel scoring function to measure the plausibility of a triple whether it is false negative or not. We assume the positive entity and the false negative entity has a substitutable relationship. The proposed false negative prediction infers whether the given negative entity can be substituted for the positive entity.

The substitution prediction can be learned together with the original loss function of knowledge graph embedding training. We do not train additional models to learn the substitution score of entities. Instead of using extra model or lots of parameters, we add only a special relation , $r_{sub}$, called “\textit{substitution}” is added to the relations set $\mathcal{R}$. The relation $r_{sub}$  will be used in calculation of substitution scores for negative entities. This special relation will be trained together with other elements in KG.

In general, many knowledge graph embedding models learn parameters using a logsigmoid loss function. Given a positive triple $(h, r ,t)$ and negative triples $(h'_{i}, r, t'_{i})$, the loss $\mathcal{L}_{KG}(\theta)$ can be expressed as follows,
\begin{align}
\label{eqn:logsigmoid}
    \mathcal{L}_{KG}(\theta) = & -\log\sigma(\gamma - f_{\theta}(h, r, t)) \nonumber \\
    & - \frac{1}{N} \sum_{i=1}^{N}\log\sigma(f_{\theta}(h'_{i}, r, t'_{i}) - \gamma),
\end{align}
where $\gamma$ is fixed margin and $\sigma$ is sigmoid function. Generally, when generating negative triples $(h'_{i}, r, t'_{i})$, some triples that already observed in the training set are filtered out so that prevent to generate false negative triples. 

However, we do not filter the false negative triples can be seen in training, but rather try to learn the pattern of false negative triples through them. We modify loss function using substitution scores which manipulate negative scores. The modified loss function can be formulated as,
\begin{align}
\label{eqn:modified_logsigmoid}
    & \mathcal{L}_{KG}(\theta) = -\log\sigma(\gamma - f_{\theta}(h, r, t)) \nonumber \\
    & \hspace{4em} - \frac{1}{N} \sum_{i=1}^{N}(1-y_{i})\log\sigma(f_{\theta}(h'_{i}, r, t'_{i}) \nonumber \\
    & \hspace{7em} - \lambda_{1} f_{\theta}(t, r_{sub}, t'_{i}) - \gamma),
\end{align}
where $f_{\theta}(t, r_{sub}, t'_{i})$ is substitution score for each negative sample, $\lambda_{1}$ is a hyperparameter for down-weighting value and $y_{i}$ is label, $1$ if the tirple is false negative and $0$ otherwise. If corrupted entity is $h'_{i}$, use $f_{\theta}(h_{i}, r_{sub}, h'_{i})$ for substitution score instead. In the modified loss, the negative scores are reduced by the substitution scores of between positive and corrupted entities. As a result, negative triples with a high substitution scores have a reduced triple scores and the effect of the negative samples on the loss are decreased.

In the false negative triples, the substitution score $f_{\theta}(t_{i}, r_{sub}, t'_{i})$ must be high because the positive entity $t$ and negative entity $t'$ can be seen substitutable. On the other hand, the influence of substitution scores of true negative samples should be minimized so that the true negatives can contribute to learning. For this, we define an auxiliary loss for substitution score prediction and train it together with the modified loss function $\mathcal{L}_{KG}$. The loss for learning the substitution score, $\mathcal{L}_{SUB}$, is also defined by using the logsigmoid loss function, written as,
\begin{equation}
\label{eqn:logsigmoid_with_aux_loss}
    \mathcal{L}_{SUB}(\theta) = - \frac{\lambda_{2}}{N}\sum_{i=1}^{N} y_{i}\log\sigma(f_{\theta}(t, r_{sub}, t'_{i})),
\end{equation}
where $\lambda_{2}$ is hyperparameter. However, following this equation, the substitution scores are trained only increasing way, thus it is not possible to calculate the correct scores. Instead of using negative samples like the original knowledge graph embedding loss, we penalize the substitution score by using an additional regularization term. The modified substitution loss with regularization term can be formalized as, 
\begin{align}
\label{eqn:logsigmoid_with_regul}
    \mathcal{L}_{SUB}(\theta) = &- \frac{\lambda_{2}}{N}\sum_{i=1}^{N} y_{i}\log\sigma(f_{\theta}(t, r_{sub}, t'_{i})) \nonumber \\
    & + \lambda_{1} \lVert \sum_{i=1}^{N}f_{\theta}(t, r_{sub}, t'_{i}) \rVert_{1}.
\end{align}

By regularizing all negative samples, we can make substitution scores of true negative converge to $0$. Through this, we can keep the effect of true negative triples and prevent entire substitution scores getting larger. We can effectively avoid the situation of using negative samples again in substitution loss to predict false negative samples by using regularization.

The final loss $\mathcal{L}$ is sum of $\mathcal{L}_{KG}$ and $\mathcal{L}_{SUB}$ as follows,
\begin{equation}
\label{eqn:final_loss}
    \mathcal{L}(\theta) = \mathcal{L}_{KG}(\theta) + \mathcal{L}_{SUB}(\theta),
\end{equation}
and embedding model are trained to optimize this loss function. We just modify objective loss function for models and do not manipulate any scoring functions. In the experiment results, we figure out that the substitution loss contributes to the training of the knowledge graph embedding.
\section{Experiments}

\begin{table}
\centering
\resizebox{\columnwidth}{!}{
\begin{tabular}{c|ccccc}
\hline
Dataset & \#entity & \#relation & \#train & \#valid & \#test \\
\hline
FB15K237 & 14,541 & 237 & 272,115 & 17,535 & 20,466 \\
WN18RR & 40,943 & 11 & 86,835 & 3,034 & 3,134 \\
\hline
\end{tabular}
}
\caption{Statistics of FB15K-237 and WN18RR datasets.}
\label{tab:data_statistics}
\end{table}

To evaluate our method, we compare the performances of five different knowledge graph embedding models on two benchmark datasets which are widely used in link prediction problem. All models and algorithms are implemented through PyTorch framework and run on a single NVIDIA V100 GPU machine with 32GB RAM.\footnote{The codes of this paper are available at \url{https://github.com/sh-je/EANS}}

\begin{table*}[h]
\centering
\resizebox{\textwidth}{!}{
\begin{tabular}{c | c | ccc | ccc}
\hline
\multirow{2}{*}{\textbf{Scoring Function}} & \multirow{2}{*}{\textbf{Sampling Method}} &  \multicolumn{3}{c|}{\textbf{FB15K237}} &  \multicolumn{3}{c}{\textbf{WN18RR}} \\
 & & \textbf{MR} & \textbf{MRR} & \textbf{Hit@10} & \textbf{MR} & \textbf{MRR} & \textbf{Hit@10} \\
\hline
\multirow{7}{*}{\shortstack{TransE \\ \cite{bordes2013translating}}} & Uniform$^{\dagger}$ & 357 & 0.294 & 0.465 & \textbf{3384} & \underline{0.226} & 0.501\\
& KBGAN\cite{cai2017kbgan}$^{\dagger\dagger}$ & 722 & 0.293 & 0.466 & 5356 & 0.181 & 0.432\\
& NSCaching\cite{zhang2019nscaching} & 186 & 0.299 & 0.476 & 4472 & 0.200 & 0.478\\
& Self-adv.\cite{sun2019rotate} & \underline{172} & 0.330 & \underline{0.526} & \underline{3429} & 0.223 & 0.530\\
& SANS + Self-adv.\cite{ahrabian2020structure} & - & 0.327 & 0.520 & - & 0.225 & \underline{0.532}\\
& EANS (ours) & \textbf{169} & \underline{0.338} & \underline{0.526} & 3488 & 0.222 & 0.526\\
& EANS + Self-adv. (ours)& \underline{172} & \textbf{0.342} & \textbf{0.534} & 3686 & \textbf{0.228} & \textbf{0.533}\\
\hline
\multirow{6}{*}{\shortstack{TransD \\ \cite{ji2015knowledge}}} & Uniform$^{\dagger\dagger}$ & \underline{188} & 0.245 & 0.429 & \underline{3555} & 0.190 & 0.464\\
& KBGAN$^{\dagger\dagger}$ & 825 & 0.247 & 0.444 & 4083 & 0.188 & 0.464\\
& NSCaching & 189 & 0.286 & 0.479 & \textbf{3104} & 0.201 & \underline{0.484}\\
& Self-adv. & \textbf{184} & \underline{0.334} & \underline{0.529} & 5520 & 0.211 & 0.477\\
& EANS (ours) & 208 & \underline{0.334} & 0.519 & 6937 & \underline{0.218} & 0.476\\
& EANS + Self-adv. (ours)& \textbf{184} & \textbf{0.340} & \textbf{0.534} & 6640 & \textbf{0.225} & \textbf{0.491}\\
\hline
\multirow{7}{*}{\shortstack{DistMult \\ \cite{yang2014embedding}}} & Uniform$^{\ddagger}$ & \underline{254} & 0.241 & 0.419 & 5110 & 0.430 & 0.490\\
& KBGAN$^{\dagger\dagger}$ & 276 & 0.227 & 0.400 & 11351 & 0.204 & 0.295\\
& NSCaching & 273 & 0.283 & 0.456 & 7708 & 0.413 & 0.455\\
& Self-adv. & \textbf{173} & \underline{0.309} & 0.484 & \textbf{4765} & \textbf{0.439} & \underline{0.536}\\
& SANS + Self-adv. & - & \textbf{0.310} & \underline{0.487} & - & 0.368 & 0.387\\
& EANS (ours) & 397 & \underline{0.309} & 0.482 & \underline{4938} & \underline{0.438} & \textbf{0.538}\\
& EANS + Self-adv. (ours)& 472 & 0.304 & \textbf{0.489} & 5584 & 0.431 & 0.518\\
\hline
\multirow{6}{*}{\shortstack{ComplEx \\ \cite{trouillon2016complex}}} & Uniform$^{\ddagger}$ & 339 & 0.247 & 0.428 & \underline{5261} & 0.440 & 0.510\\
& KBGAN$^{\dagger\dagger}$ & 881 & 0.191 & 0.321 & 7528 & 0.318 & 0.355\\
& NSCaching & \underline{221} & 0.302 & 0.481 & 5365 & 0.446 & 0.509\\
& Self-adv. & \textbf{166} & \underline{0.322} & \textbf{0.512} & \textbf{5226} & \textbf{0.468} & \textbf{0.558}\\
& EANS (ours) & 454 & \textbf{0.323} & \underline{0.503} & 5350 & \underline{0.463} & \textbf{0.558}\\
& EANS + Self-adv. (ours) & 446 & 0.292 & 0.480 & 6709 & 0.456 & \underline{0.532}\\
\hline
\multirow{5}{*}{\shortstack{RotatE \\ \cite{sun2019rotate}}} & Uniform & 187 & 0.295 & 0.478 & \underline{3274} & 0.473 & 0.565\\
& Self-adv. & 177 & 0.338 & \underline{0.533} & 3340 & 0.476 & 0.571\\
& SANS + Self-adv. & - & 0.336 & 0.531 & - & 0.475 & 0.571\\
& EANS (ours) & \underline{169} & \underline{0.341} & 0.528 & \textbf{3149} & \underline{0.487} & \underline{0.574}\\
& EANS + Self-adv. (ours)& \textbf{165} & \textbf{0.344} & \textbf{0.537} & 3402 & \textbf{0.489} & \textbf{0.576}\\
\hline
\end{tabular}
}
\caption{Comparison of different negative sampling methods on FB15K-237 and WN18RR. Results of [$^{\dagger}$] are taken from \cite{nguyen2017novel}, [$^{\ddagger}$] are from\cite{dettmers2018convolutional}, [$^{\dagger\dagger}$] are from \cite{zhang2019nscaching}, and the other results are taken from the corresponding papers. Bold numbers represent the best and underlined numbers represent the second best.}
\label{tab:main_result}
\end{table*}

\subsection{Experiment Settings}
\subsubsection{Hyperparameter Settings}
We use Adam \cite{kingma2014adam} to optimize the experimental models and methods. The hyperparameters are validated in the following range, embedding dimension $d \in \{100, 200, 1000\}$, mini-batch size $b \in \{512, 1024, 2048\}$, fixed margin $\gamma \in \{6, 9, 12 , 18, 24\}$, regularization weight $\lambda_{1} \in \{0.1, 0.05, 0.01, 0.001\}$, and substitution loss weight $\lambda_{2} \in \{1.0, 0.5, 0.1, 0.05\}$. The number of clusters for K-means $k$ is fixed to $100$ and Gaussian variance $\sigma$ is set to $\frac{2|\mathcal{E}|}{k}$. All models train 100,000 steps on FB15K-237 and 80,000 steps on WN18RR. The embeddings are re-ordered every 1000 step in training. The optimal hyperparameters settings for EANS are summarized in Table \ref{tab:hyperparams_fb15k237} and \ref{tab:hyperparams_wn18rr} in Appendix.

\subsubsection{Datasets and Evaluation Metrics}

We evaluate on two datasets, FB15K-237 \cite{toutanova2015observed} and WN18RR \cite{dettmers2018convolutional}. FB15K-237 and WN18RR are subsets of FB15K \cite{bordes2013translating, bollacker2008freebase} and WN18 \cite{bordes2013translating, miller1995wordnet}, respectively. In FB15K-237 and WN18RR, some relations that can be easily inferred from inverse-relation have been removed. Since these datasets contain more realistic and refined triples than original datasets, the performance of the model on these datasets can be compared more meaningful. Some statistics of datasets are summarized in Table \ref{tab:data_statistics}. 

In evaluating our methods, we use the standard evaluation metrics mean rank(MR), mean reciprocal rank(MRR), and hits at N(Hit@N). We measure the evaluation metric for all results with filtered setting which is same as \cite{bordes2013translating}.

\subsection{Main Results}
\begin{figure*}[h]
    \centering
    \includegraphics[width=\textwidth]{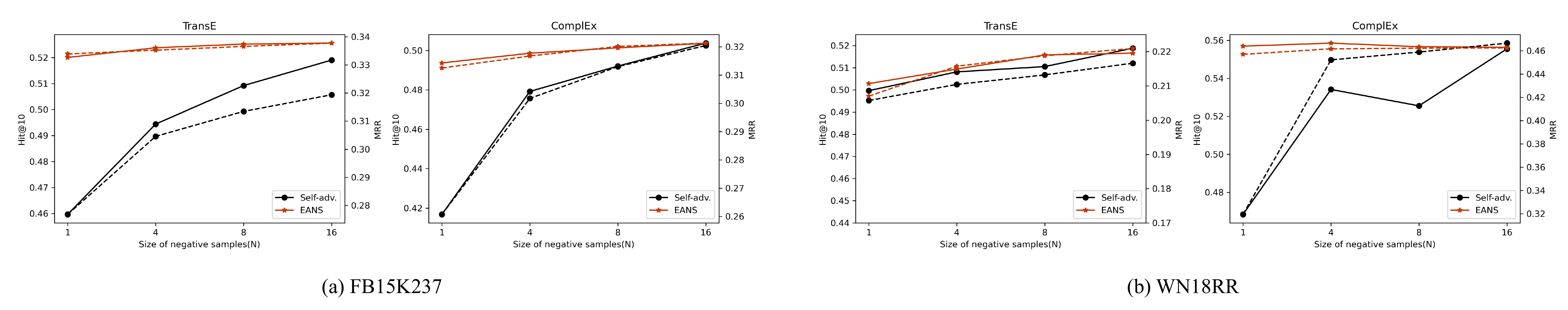}
    \caption{Hit@10(solid line) and MRR(dashed line) of TransE and ComplEx with different negative sample sizes.}
    \label{fig:small_ns_results}
\end{figure*}
We apply our EANS to five different models and compare them with several state-of-the-art negative sampling methods. The models used for comparison are TransE\cite{bordes2013translating}, TransD\cite{ji2015knowledge}, DistMult\cite{yang2014embedding}, ComplEx\cite{trouillon2016complex}, and RotatE\cite{sun2019rotate}. 
We compare the performances of the models with KB-GAN\cite{cai2017kbgan}, NS-Caching\cite{zhang2019nscaching}, self-adversarial sampling\cite{sun2019rotate}, SANS\cite{ahrabian2020structure} and our EANS methods. Our EANS method can be used with self-adversarial negative sampling together, and the performances are also confirmed through experiments. Table \ref{tab:main_result} shows the results of our experiments.

EANS always outperforms all the state-of-the-art methods when combined with translational models(TransE, TransD). RotatE with EANS also achieves the best performance on both two datasets. We find that our methods when combined with semantic matching score models(DistMult, ComplEx) are not the best in MRR, but the second best performance. The reason why our mehtods show those difference may be because the entity clustering process in EANS is based on distance metric.

\subsection{Small-Sized Negative Sampling Result}
We evaluate our method with extremely reduced the negative samples. Table \ref{tab:one_ns_result} shows the performance when EANS negative sample size is applied as 1. Even though only one negative is used, they EANS is only different about 1-2\% on Hit@10 from the best performance of the other state-of-the-art methods. Especially, ComplEx in WN18RR, the performance of only-one-negative setting is only 0.1\% different on Hit@10 from the best performance. The results of other scoring function models with small negative samples are reported in Appendix.

We also apply small negative samples to the self-adversarial negative sampling, which is one of the most effective sampling methods. Figure \ref{fig:small_ns_results} shows the results of TransE and ComplEx with negative sample size $ n \in \{1,4,8,16\}$. In the results, performances of self-adversarial sampling drastically decrease as the negative sample size goes down, but EANS usually does not.

\begin{table}[h]
\centering
\resizebox{\columnwidth}{!}{
\begin{tabular}{c | ccc | ccc}
\hline
\multirow{2}{*}{\shortstack{\textbf{Scoring} \\ \textbf{Function}}} &  \multicolumn{3}{c|}{\textbf{FB15K237}} &  \multicolumn{3}{c}{\textbf{WN18RR}} \\
 & \textbf{MR} & \textbf{MRR} & \textbf{H@10} & \textbf{MR} & \textbf{MRR} & \textbf{H@10} \\
\hline
TransE & 179 & .334 & .520 & 4157 & .207 & .503\\
\hline
TransD & 213 & .328 & .510 & 6409 & .216 & .478\\
\hline
DistMult & 418 & .294 & .467 & 4933 & .433 & .530\\
\hline
ComplEx & 491 & .312 & .494 & 5202 & .457 & .557\\
\hline
RotatE & 165 & .328 & .517 & 3741 & .474 & .559\\
\hline
\end{tabular}
}
\caption{Performances of EANS with only one negative sample(n=1). Even though only one negative is used, they show comparable results to the other methods.}
\label{tab:one_ns_result}
\end{table}

\begin{table}[h]
\centering
\begin{tabular}{c | cc | cc}
\hline
\multirow{2}{*}{\textbf{Step}} &  \multicolumn{2}{c|}{\textbf{Uniform}} &  \multicolumn{2}{c}{\textbf{EANS}} \\
 & \textbf{pos.} & \textbf{neg.} & \textbf{pos.} & \textbf{neg.} \\
\hline
1K &  -0.516 & -0.638 & -0.467 & -0.393 \\
10K &  -0.256 & -4.455 & -0.299 & -1.709 \\
100K &  -0.258 & -5.805 & -0.275 & -1.856 \\
\hline
\end{tabular}
\caption{Average scores of positive and negative samples of various steps on uniform sampling method and EANS with TransE on FB15K-237.}
\label{tab:pos_neg_comparison}
\end{table}

\subsection{Quality of Negative Samples from EANS}
We compare the negative triple scores calculated by TransE in the training process on the FB15K-237 dataset using EANS and uniform sampling. We check the scores of negative samples in each 1k, 10k, and 100k training step.  Table \ref{tab:pos_neg_comparison} shows the average scores of the positive and negative triples in 1,000 mini-batches calculated by TransE. 

As training progresses, the average score of negative triples in the uniform sampling method gradually decreases, and gap between positives and negatives increases. Although the scale of the scores calculated by both methods is similar, the average of negative scores by EANS is higher than the uniform sampling method's scores. It can be seen that EANS generates a large number of high-quality negative samples than uniform negative sampling.

\begin{figure}[h]
    \centering
    \includegraphics[width=\columnwidth]{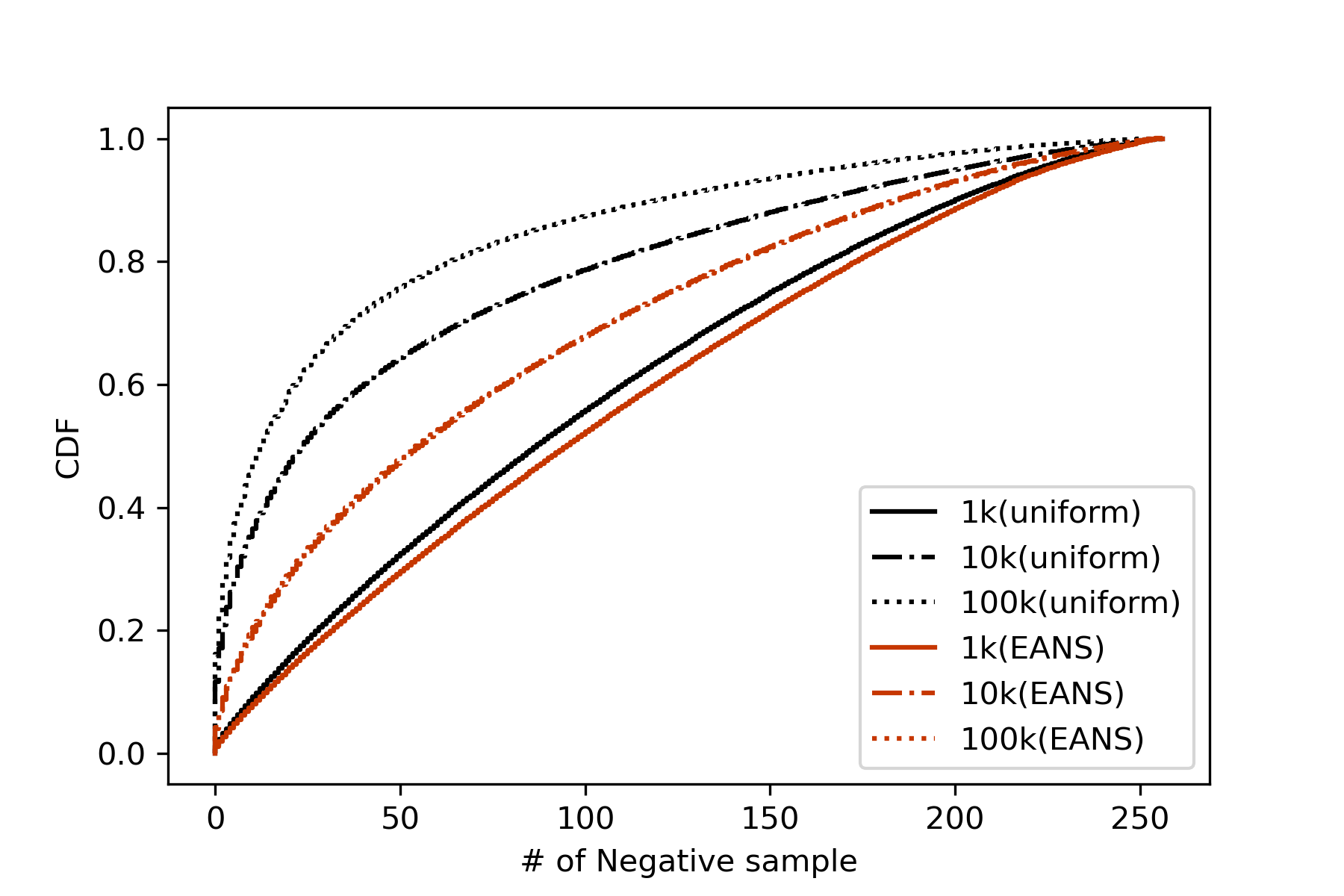}
    \caption{Cumulative density functions of sorted negative scores for each positive sample from TransE on FB15K237.}
    \label{fig:neg_probs_cdf}
\end{figure}

Additionally, we check the distribution of negative scores for each positive sample by applying the softmax function. The scores are averaged in a batch and sorted in descending order. Through Figure \ref{fig:neg_probs_cdf}, we can find that density of the uniform sampling method is skewed to a few top ranked negative samples over time. When training reaches 100K, the only top-64 negative samples account for more than 80\% of the total weight, while the distribution of EANS is quite even.

\subsection{Effectiveness of Substitution Scores for False Negative Prediction}

\begin{figure}[h]
    \centering
    \includegraphics[width=\columnwidth]{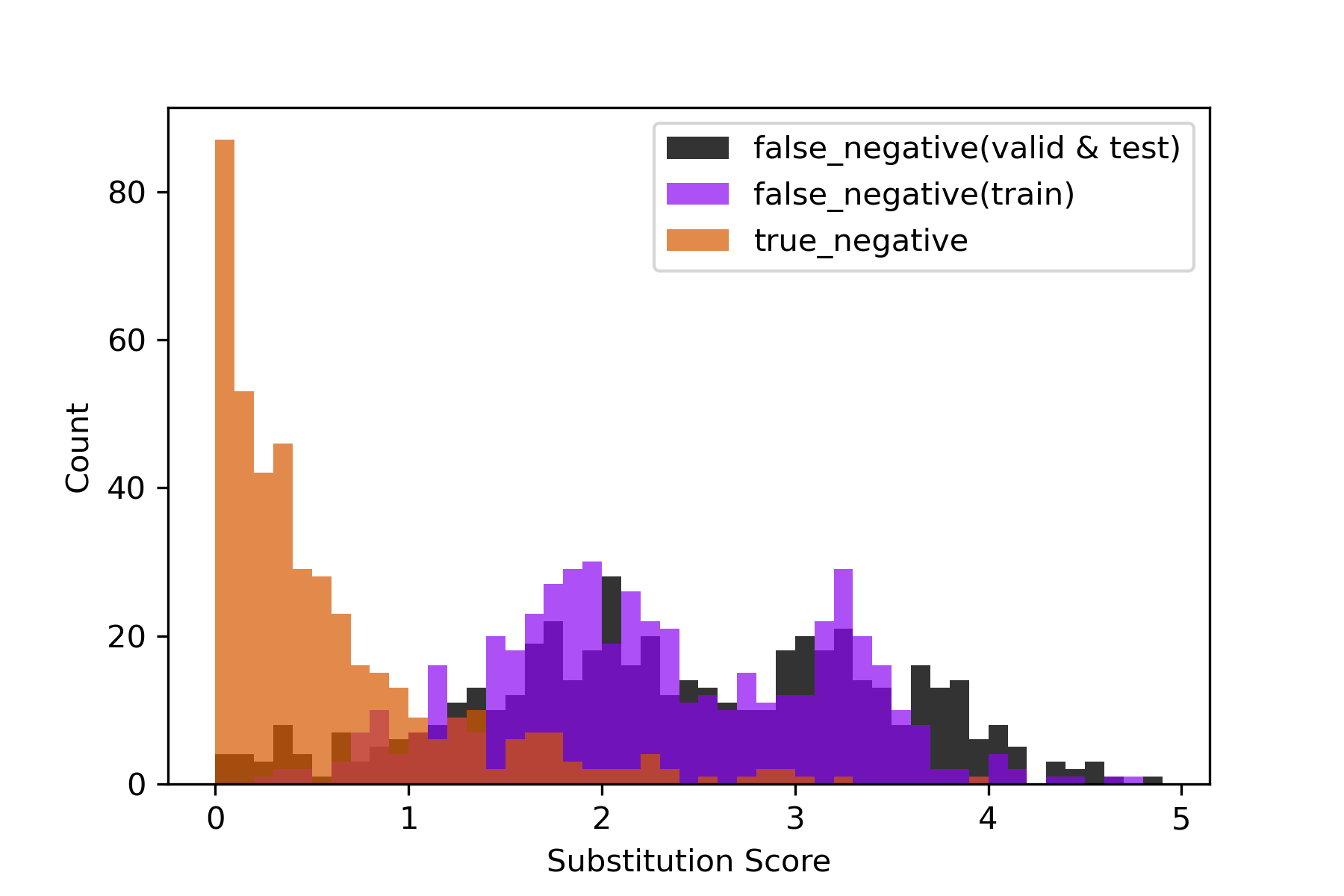}
    \caption{Histogram of substitution scores of negative samples from TransE on FB15K-237. The false negative triples can be observed in train set(purple) or evaluation set(black).}
    \label{fig:subs_hist}
\end{figure}

With trained TransE on the FB15K-237, we divide the negative samples into three groups according to their truth values, and check the substitution score distribution in each group. The histogram of substitution scores in each group is depicted in Figure \ref{fig:subs_hist}. We find that proposed substitution scores can be used to discriminate the true and false negatives. Although the false negative triples which can be observed in evaluation sets are not used in training, they also have high substitution scores. This result shows that the learned \textit{substitution} relation can infer the substitution relationship between entities which has not seen on training.

Additionally, we do ablation test to check whether false negative prediction is helpful for learning. We evaluate EANS combined with RotatE, which showed the best performance among the various models. The two parts of EANS, that extracting entity aware negative from Gaussian distribution and predicting false negatives with substitution loss are separatly adopted. Table \ref{tab:ablation} shows that separated methods can not produce good performance as performance of whole EANS method. The part using Gaussian sampling (\textit{Gauss.}) improves performance on FB15K-237, but not on WN18RR. On the other hand, the part using substitution loss (\textit{Subs.}) shows good performance on WN18RR, but not on FB15K-237.

\begin{table}[h]
\centering
\resizebox{\columnwidth}{!}{
\begin{tabular}{cc | cc | cc}
\hline
\multicolumn{2}{c|}{\textbf{Ablation}} & \multicolumn{2}{c|}{\textbf{FB15K237}} &  \multicolumn{2}{c}{\textbf{WN18RR}} \\ 
\textbf{Gauss.} & \textbf{Subs.} & \textbf{MRR} & \textbf{H@10} & \textbf{MRR} & \textbf{H@10}\\
\hline
\xmark & \xmark & 0.295 & 0.478 & 0.473 & 0.565 \\
\cmark & \xmark & 0.318 & 0.514 & 0.470 & 0.560 \\
\xmark & \cmark & 0.279 & 0.449 & 0.485 & 0.573 \\
\cmark & \cmark & \textbf{0.341} & \textbf{0.528} & \textbf{0.487} & \textbf{0.574} \\
\hline
\end{tabular}
}
\caption{Performance of EANS combined with RotatE in ablation settings. 'Gauss.' represents adopting Gaussian distribution for entity sampling and 'Subs.' represents adopting an auxiliary substitution loss.}
\label{tab:ablation}
\end{table}

\section{Conclusions}

We propose a novel negative sampling method, EANS, which can sample high-quality negatives based on the given positive entity. The proposed method samples hard negative entities by utilizing entity clustering and Gaussian distribution, and effectively suppresses the influence of false negatives by optimizing the additional false negative prediction loss. Through various analyses, we confirm that each component of EANS contributes to generate high-quality negative samples. Our experimental results show that the performances of proposed EANS combined with several knowledge graph embedding models outperform existing the state-of-the-art negative sampling methods on two standard benchmarks. Moreover, EANS also achieves competitive performance even when the size of the negative sample used for learning is limited to only one. 
\section{Limitations}

The biggest limitation of EANS is that a clustering algorithm must be applied to align entire entities. In EANS, it is necessary to cluster and re-order all entity embeddings every 1-3 epochs using the clustering algorithm. Without clustering process, entity aware negatives cannot be sampled. In this paper, we applied the K-means algorithm with complexity dependent on the size of entities set and entity embedding dimension. Even if entity clustering is applied intermittently in EANS, if the size of entities set increases dramatically or the size of entity embedding dimension becomes extremely large, the entity clustering process may become a bottleneck. Therefore, a more efficient entity alignment method may be needed to apply to large scale KG. In the future, we plan to study methods to estimate entity distribution more efficiently and effectively to overcome this limitation. 

\bibliography{custom}

\begin{thebibliography}{33}
\expandafter\ifx\csname natexlab\endcsname\relax\def\natexlab#1{#1}\fi

\bibitem[{Ahrabian et~al.(2020)Ahrabian, Feizi, Salehi, Hamilton, and
  Bose}]{ahrabian2020structure}
Kian Ahrabian, Aarash Feizi, Yasmin Salehi, William~L Hamilton, and
  Avishek~Joey Bose. 2020.
\newblock Structure aware negative sampling in knowledge graphs.
\newblock \emph{arXiv preprint arXiv:2009.11355}.

\bibitem[{Auer et~al.(2007)Auer, Bizer, Kobilarov, Lehmann, Cyganiak, and
  Ives}]{auer2007dbpedia}
S{\"o}ren Auer, Christian Bizer, Georgi Kobilarov, Jens Lehmann, Richard
  Cyganiak, and Zachary Ives. 2007.
\newblock Dbpedia: A nucleus for a web of open data.
\newblock In \emph{The semantic web}, pages 722--735. Springer.

\bibitem[{Bollacker et~al.(2008)Bollacker, Evans, Paritosh, Sturge, and
  Taylor}]{bollacker2008freebase}
Kurt Bollacker, Colin Evans, Praveen Paritosh, Tim Sturge, and Jamie Taylor.
  2008.
\newblock Freebase: a collaboratively created graph database for structuring
  human knowledge.
\newblock In \emph{Proceedings of the 2008 ACM SIGMOD international conference
  on Management of data}, pages 1247--1250.

\bibitem[{Bordes et~al.(2013)Bordes, Usunier, Garcia-Duran, Weston, and
  Yakhnenko}]{bordes2013translating}
Antoine Bordes, Nicolas Usunier, Alberto Garcia-Duran, Jason Weston, and Oksana
  Yakhnenko. 2013.
\newblock Translating embeddings for modeling multi-relational data.
\newblock \emph{Advances in neural information processing systems}, 26.

\bibitem[{Cai and Wang(2017)}]{cai2017kbgan}
Liwei Cai and William~Yang Wang. 2017.
\newblock Kbgan: Adversarial learning for knowledge graph embeddings.
\newblock \emph{arXiv preprint arXiv:1711.04071}.

\bibitem[{Dettmers et~al.(2018)Dettmers, Minervini, Stenetorp, and
  Riedel}]{dettmers2018convolutional}
Tim Dettmers, Pasquale Minervini, Pontus Stenetorp, and Sebastian Riedel. 2018.
\newblock Convolutional 2d knowledge graph embeddings.
\newblock In \emph{Proceedings of the AAAI Conference on Artificial
  Intelligence}, volume~32.

\bibitem[{Goodfellow et~al.(2014)Goodfellow, Pouget-Abadie, Mirza, Xu,
  Warde-Farley, Ozair, Courville, and Bengio}]{goodfellow2014generative}
Ian Goodfellow, Jean Pouget-Abadie, Mehdi Mirza, Bing Xu, David Warde-Farley,
  Sherjil Ozair, Aaron Courville, and Yoshua Bengio. 2014.
\newblock Generative adversarial nets.
\newblock \emph{Advances in neural information processing systems}, 27.

\bibitem[{Hajimoradlou and Kazemi(2022)}]{hajimoradlou2022stay}
Ainaz Hajimoradlou and Mehran Kazemi. 2022.
\newblock Stay positive: Knowledge graph embedding without negative sampling.
\newblock \emph{arXiv preprint arXiv:2201.02661}.

\bibitem[{Huang et~al.(2019)Huang, Zhang, Li, and Li}]{huang2019knowledge}
Xiao Huang, Jingyuan Zhang, Dingcheng Li, and Ping Li. 2019.
\newblock Knowledge graph embedding based question answering.
\newblock In \emph{Proceedings of the twelfth ACM international conference on
  web search and data mining}, pages 105--113.

\bibitem[{Ji et~al.(2015)Ji, He, Xu, Liu, and Zhao}]{ji2015knowledge}
Guoliang Ji, Shizhu He, Liheng Xu, Kang Liu, and Jun Zhao. 2015.
\newblock Knowledge graph embedding via dynamic mapping matrix.
\newblock In \emph{Proceedings of the 53rd annual meeting of the association
  for computational linguistics and the 7th international joint conference on
  natural language processing (volume 1: Long papers)}, pages 687--696.

\bibitem[{Kingma and Ba(2014)}]{kingma2014adam}
Diederik~P Kingma and Jimmy Ba. 2014.
\newblock Adam: A method for stochastic optimization.
\newblock \emph{arXiv preprint arXiv:1412.6980}.

\bibitem[{Lin et~al.(2015)Lin, Liu, Sun, Liu, and Zhu}]{lin2015learning}
Yankai Lin, Zhiyuan Liu, Maosong Sun, Yang Liu, and Xuan Zhu. 2015.
\newblock Learning entity and relation embeddings for knowledge graph
  completion.
\newblock In \emph{Twenty-ninth AAAI conference on artificial intelligence}.

\bibitem[{Liu et~al.(2021)Liu, Wan, He, Peng, and Yu}]{liu2021kg}
Ye~Liu, Yao Wan, Lifang He, Hao Peng, and Philip~S Yu. 2021.
\newblock Kg-bart: Knowledge graph-augmented bart for generative commonsense
  reasoning.
\newblock In \emph{Proceedings of the AAAI Conference on Artificial
  Intelligence}, volume~35, pages 6418--6425.

\bibitem[{Lloyd(1982)}]{lloyd1982least}
Stuart Lloyd. 1982.
\newblock Least squares quantization in pcm.
\newblock \emph{IEEE transactions on information theory}, 28(2):129--137.

\bibitem[{Lu et~al.(2022)Lu, Hu, and Lin}]{lu2022dense}
Haonan Lu, Hailin Hu, and Xiaodong Lin. 2022.
\newblock Dense: An enhanced non-commutative representation for knowledge graph
  embedding with adaptive semantic hierarchy.
\newblock \emph{Neurocomputing}.

\bibitem[{Miller(1995)}]{miller1995wordnet}
George~A Miller. 1995.
\newblock Wordnet: a lexical database for english.
\newblock \emph{Communications of the ACM}, 38(11):39--41.

\bibitem[{Nguyen et~al.(2017)Nguyen, Nguyen, Nguyen, and
  Phung}]{nguyen2017novel}
Dai~Quoc Nguyen, Tu~Dinh Nguyen, Dat~Quoc Nguyen, and Dinh Phung. 2017.
\newblock A novel embedding model for knowledge base completion based on
  convolutional neural network.
\newblock \emph{arXiv preprint arXiv:1712.02121}.

\bibitem[{Nickel et~al.(2011)Nickel, Tresp, and Kriegel}]{nickel2011three}
Maximilian Nickel, Volker Tresp, and Hans-Peter Kriegel. 2011.
\newblock A three-way model for collective learning on multi-relational data.
\newblock In \emph{Icml}.

\bibitem[{Saxena et~al.(2020)Saxena, Tripathi, and
  Talukdar}]{saxena2020improving}
Apoorv Saxena, Aditay Tripathi, and Partha Talukdar. 2020.
\newblock Improving multi-hop question answering over knowledge graphs using
  knowledge base embeddings.
\newblock In \emph{Proceedings of the 58th annual meeting of the association
  for computational linguistics}, pages 4498--4507.

\bibitem[{Suchanek et~al.(2007)Suchanek, Kasneci, and
  Weikum}]{suchanek2007yago}
Fabian~M Suchanek, Gjergji Kasneci, and Gerhard Weikum. 2007.
\newblock Yago: a core of semantic knowledge.
\newblock In \emph{Proceedings of the 16th international conference on World
  Wide Web}, pages 697--706.

\bibitem[{Sun et~al.(2019)Sun, Deng, Nie, and Tang}]{sun2019rotate}
Zhiqing Sun, Zhi-Hong Deng, Jian-Yun Nie, and Jian Tang. 2019.
\newblock Rotate: Knowledge graph embedding by relational rotation in complex
  space.
\newblock \emph{arXiv preprint arXiv:1902.10197}.

\bibitem[{Toutanova and Chen(2015)}]{toutanova2015observed}
Kristina Toutanova and Danqi Chen. 2015.
\newblock Observed versus latent features for knowledge base and text
  inference.
\newblock In \emph{Proceedings of the 3rd workshop on continuous vector space
  models and their compositionality}, pages 57--66.

\bibitem[{Trouillon et~al.(2016)Trouillon, Welbl, Riedel, Gaussier, and
  Bouchard}]{trouillon2016complex}
Th{\'e}o Trouillon, Johannes Welbl, Sebastian Riedel, {\'E}ric Gaussier, and
  Guillaume Bouchard. 2016.
\newblock Complex embeddings for simple link prediction.
\newblock In \emph{International conference on machine learning}, pages
  2071--2080. PMLR.

\bibitem[{Vashishth et~al.(2020)Vashishth, Sanyal, Nitin, Agrawal, and
  Talukdar}]{vashishth2020interacte}
Shikhar Vashishth, Soumya Sanyal, Vikram Nitin, Nilesh Agrawal, and Partha
  Talukdar. 2020.
\newblock Interacte: Improving convolution-based knowledge graph embeddings by
  increasing feature interactions.
\newblock In \emph{Proceedings of the AAAI conference on artificial
  intelligence}, volume~34, pages 3009--3016.

\bibitem[{Vashishth et~al.(2019)Vashishth, Sanyal, Nitin, and
  Talukdar}]{vashishth2019composition}
Shikhar Vashishth, Soumya Sanyal, Vikram Nitin, and Partha Talukdar. 2019.
\newblock Composition-based multi-relational graph convolutional networks.
\newblock \emph{arXiv preprint arXiv:1911.03082}.

\bibitem[{Wang et~al.(2018)Wang, Li, and Pan}]{wang2018incorporating}
Peifeng Wang, Shuangyin Li, and Rong Pan. 2018.
\newblock Incorporating gan for negative sampling in knowledge representation
  learning.
\newblock In \emph{Proceedings of the AAAI Conference on Artificial
  Intelligence}, volume~32.

\bibitem[{Wang et~al.(2019)Wang, He, Cao, Liu, and Chua}]{wang2019kgat}
Xiang Wang, Xiangnan He, Yixin Cao, Meng Liu, and Tat-Seng Chua. 2019.
\newblock Kgat: Knowledge graph attention network for recommendation.
\newblock In \emph{Proceedings of the 25th ACM SIGKDD international conference
  on knowledge discovery \& data mining}, pages 950--958.

\bibitem[{Wang et~al.(2014)Wang, Zhang, Feng, and Chen}]{wang2014knowledge}
Zhen Wang, Jianwen Zhang, Jianlin Feng, and Zheng Chen. 2014.
\newblock Knowledge graph embedding by translating on hyperplanes.
\newblock In \emph{Proceedings of the AAAI Conference on Artificial
  Intelligence}, volume~28.

\bibitem[{Wu et~al.(2020)Wu, Li, Zhang, Zhou, and Wu}]{wu2020diverse}
Sixing Wu, Ying Li, Dawei Zhang, Yang Zhou, and Zhonghai Wu. 2020.
\newblock Diverse and informative dialogue generation with context-specific
  commonsense knowledge awareness.
\newblock In \emph{Proceedings of the 58th annual meeting of the association
  for computational linguistics}, pages 5811--5820.

\bibitem[{Xu et~al.(2020)Xu, Ruan, Korpeoglu, Kumar, and Achan}]{xu2020product}
Da~Xu, Chuanwei Ruan, Evren Korpeoglu, Sushant Kumar, and Kannan Achan. 2020.
\newblock Product knowledge graph embedding for e-commerce.
\newblock In \emph{Proceedings of the 13th international conference on web
  search and data mining}, pages 672--680.

\bibitem[{Yang et~al.(2014)Yang, Yih, He, Gao, and Deng}]{yang2014embedding}
Bishan Yang, Wen-tau Yih, Xiaodong He, Jianfeng Gao, and Li~Deng. 2014.
\newblock Embedding entities and relations for learning and inference in
  knowledge bases.
\newblock \emph{arXiv preprint arXiv:1412.6575}.

\bibitem[{Zhang et~al.(2018)Zhang, Ai, Chen, and Wang}]{zhang2018learning}
Yongfeng Zhang, Qingyao Ai, Xu~Chen, and Pengfei Wang. 2018.
\newblock Learning over knowledge-base embeddings for recommendation.
\newblock \emph{arXiv preprint arXiv:1803.06540}.

\bibitem[{Zhang et~al.(2019)Zhang, Yao, Shao, and Chen}]{zhang2019nscaching}
Yongqi Zhang, Quanming Yao, Yingxia Shao, and Lei Chen. 2019.
\newblock Nscaching: simple and efficient negative sampling for knowledge graph
  embedding.
\newblock In \emph{2019 IEEE 35th International Conference on Data Engineering
  (ICDE)}, pages 614--625. IEEE.

\end{thebibliography}
\bibliographystyle{acl_natbib}

\appendix 
\section{Appendices}

\subsection{Hyperparameters}
We list the best hyperparameters of EANS on the benchmarks. Table \ref{tab:hyperparams_fb15k237} summarizes the best hyperparameters on FB15K-237 and Table \ref{tab:hyperparams_wn18rr} summarized the best hyperparameters on WN18RR. In the tables, $d$ is embedding dimension, $b$ is batch size, $n$ is negative sample size, lr is initial learning rate, $\gamma$ is fixed margin, and $\alpha$ is sampling weight when using self-adversarial sampling.

\begin{table}[h]
\centering
\resizebox{\columnwidth}{!}{
\begin{tabular}{c | c | c | c | c | c }
\hline
 & \textbf{TransE} & \textbf{TransD} & \textbf{DistMult} & \textbf{ComplEx} & \textbf{RotatE}\\
\hline
$d$ & 1000 & 1000 & 1000 & 1000 & 1000 \\
$b$ & 1024 & 1024 & 1024 & 1024 & 1024 \\
$n$ & 256 & 256 & 256 & 256 & 256 \\
lr & 5e-5 & 5e-5 & 0.001 & 0.001 & 5e-5\\
$\gamma$ & 9.0 & 9.0 & 200.0 & 200.0 & 9.0 \\
$\alpha$ & 1.0 & 1.0 & 1.0 & 1.0 & 1.0 \\ 
$\lambda_{1}$ & 0.1 & 0.1 & 0.05 & 0.05 & 0.1 \\
$\lambda_{2}$ & 1.0 & 1.0 & 1.0 & 1.0 & 1.0 \\
\hline
\end{tabular}
}
\caption{The best hyperparameter settings of EANS on FB15K-237}
\label{tab:hyperparams_fb15k237}
\end{table}

\begin{table}[h]
\centering
\resizebox{\columnwidth}{!}{
\begin{tabular}{c | c | c| c | c | c }
\hline
 & \textbf{TransE} & \textbf{TransD} & \textbf{DistMult} & \textbf{ComplEx} & \textbf{RotatE}\\
\hline
$d$ & 500 & 500 & 500 & 500 & 500 \\
$b$ & 512 & 512 & 512 & 512 & 512 \\
$n$ & 1024 & 1024 & 1024 & 1024 & 1024 \\
lr & 5e-5 & 5e-5 & 0.002 & 0.002 & 5e-5\\
$\gamma$ & 6.0 & 6.0 & 200.0 & 200.0 & 6.0 \\
$\alpha$ & 0.5 & 0.5 & 1.0 & 1.0 & 0.5 \\ 
$\lambda_{1}$ & 0.01 & 0.01 & 0.01 & 0.01 & 0.01 \\
$\lambda_{2}$ & 0.05 & 0.05 & 0.05 & 0.05 & 0.05 \\
\hline
\end{tabular}
}
\caption{The best hyperparameter settings of EANS on WN18RR}
\label{tab:hyperparams_wn18rr}
\end{table}

\subsection{Variance of Main Results}
Table \ref{tab:result_mean_var} shows the variance of the MRR and Hit@10 results on the two benchmarks. Both mean and standard error values are calculated by three runs of each knowledge graph embedding model with different random initialization.

\begin{table*}[h]
\centering
\resizebox{\textwidth}{!}{
\begin{tabular}{c | c | cc | cc}
\hline
\multirow{2}{*}{\textbf{Scoring Function}} & \multirow{2}{*}{\textbf{Sampling Method}} &  \multicolumn{2}{c|}{\textbf{FB15K237}} &  \multicolumn{2}{c}{\textbf{WN18RR}} \\
 & & \textbf{MRR} & \textbf{Hit@10} & \textbf{MRR} & \textbf{Hit@10} \\
\hline
\multirow{2}{*}{\shortstack{TransE \\ \cite{bordes2013translating}}} & EANS & 0.338 $\pm$ 0.000 & 0.526 $\pm$ 0.000 & 0.222 $\pm$ 0.002  & 0.526 $\pm$ 0.002 \\
& EANS + Self-adv. & 0.342 $\pm$ 0.001  & 0.534 $\pm$ 0.000 & 0.228 $\pm$ 0.003  & 0.533 $\pm$ 0.003 \\
\hline
\multirow{2}{*}{\shortstack{TransD \\ \cite{ji2015knowledge}}} & EANS & 0.334 $\pm$ 0.000 & 0.519 $\pm$ 0.000 & 0.218 $\pm$ 0.002  & 0.476 $\pm$ 0.003 \\
& EANS + Self-adv. & 0.340 $\pm$ 0.002  & 0.534 $\pm$ 0.001 & 0.225 $\pm$ 0.002  & 0.491 $\pm$ 0.003 \\
\hline
\multirow{2}{*}{\shortstack{DistMult \\ \cite{yang2014embedding}}} & EANS & 0.309 $\pm$ 0.001 & 0.482 $\pm$ 0.001 & 0.438 $\pm$ 0.001 & 0.538 $\pm$ 0.001 \\
& EANS + Self-adv. & 0.304 $\pm$ 0.000  & 0.489 $\pm$ 0.003 & 0.431 $\pm$ 0.001  & 0.518 $\pm$ 0.001 \\
\hline
\multirow{2}{*}{\shortstack{ComplEx \\ \cite{trouillon2016complex}}} & EANS & 0.323 $\pm$ 0.000 & 0.503 $\pm$ 0.001 & 0.463 $\pm$ 0.001  & 0.558 $\pm$ 0.001 \\
& EANS + Self-adv. & 0.292 $\pm$ 0.000  & 0.480 $\pm$ 0.002 & 0.456 $\pm$ 0.001  & 0.532 $\pm$ 0.002 \\
\hline
\multirow{2}{*}{\shortstack{RotatE \\ \cite{sun2019rotate}}} & EANS & 0.341 $\pm$ 0.002 & 0.528 $\pm$ 0.002 & 0.487 $\pm$ 0.001  & 0.574 $\pm$ 0.000 \\
& EANS + Self-adv. & 0.344 $\pm$ 0.001  & 0.537 $\pm$ 0.000 & 0.489 $\pm$ 0.001  & 0.576 $\pm$ 0.001 \\
\hline
\end{tabular}
}
\caption{The mean and standard error of the MRR and Hit@10 results on FB15K-237 and WN18RR}
\label{tab:result_mean_var}
\end{table*}

\subsection{Results of Different Small Negative Sample Sizes}

\begin{table*}[h!]
\centering
\resizebox{\textwidth}{!}{
\begin{tabular}{c | cc | cc | cc | cc | cc }
\hline
\textbf{Size of} & \multicolumn{2}{c|}{\textbf{TransE}} & \multicolumn{2}{c|}{\textbf{TransD}} & \multicolumn{2}{c|}{\textbf{DistMult}} & \multicolumn{2}{c|}{\textbf{ComplEx}} & \multicolumn{2}{c}{\textbf{RotatE}} \\ 
\textbf{Negative Samples} & \textbf{MRR} & \textbf{Hit@10} & \textbf{MRR} & \textbf{Hit@10} & \textbf{MRR} & \textbf{Hit@10} & \textbf{MRR} & \textbf{Hit@10} & \textbf{MRR} & \textbf{Hit@10} \\
\hline
1 & 0.334 & 0.520 & 0.328 & 0.510 & 0.294 & 0.467 & 0.312 & 0.494 & 0.328 & 0.517 \\
4 & 0.335 & 0.523 & 0.330 & 0.510 & 0.298 & 0.472 & 0.317 & 0.499 & 0.337 & 0.527 \\
8 & 0.337 & 0.525 & 0.332 & 0.516 & 0.302 & 0.476 & 0.320 & 0.501 & 0.338 & 0.527 \\
16 & 0.338 & 0.526 & 0.331 & 0.515 & 0.302 & 0.475 & 0.321 & 0.504 & 0.338 & 0.528 \\
\hline
\end{tabular}
}
\caption{The MRR and Hit@10 results of EANS with small negative samples setting on FB15K-237}
\label{tab:small_ns_fb15k237}
\end{table*}

\begin{table*}[h!]
\centering
\resizebox{\textwidth}{!}{
\begin{tabular}{c | cc | cc | cc | cc | cc }
\hline
\textbf{Size of} & \multicolumn{2}{c|}{\textbf{TransE}} & \multicolumn{2}{c|}{\textbf{TransD}} & \multicolumn{2}{c|}{\textbf{DistMult}} & \multicolumn{2}{c|}{\textbf{ComplEx}} & \multicolumn{2}{c}{\textbf{RotatE}} \\ 
\textbf{Negative Samples} & \textbf{MRR} & \textbf{Hit@10} & \textbf{MRR} & \textbf{Hit@10} & \textbf{MRR} & \textbf{Hit@10} & \textbf{MRR} & \textbf{Hit@10} & \textbf{MRR} & \textbf{Hit@10} \\
\hline
1 & 0.207 & 0.503 & 0.216 & 0.478 & 0.433 & 0.530 & 0.457 & 0.557 & 0.474 & 0.559 \\
4 & 0.215 & 0.509 & 0.213 & 0.464 & 0.437 & 0.534 & 0.462 & 0.559 & 0.473 & 0.561 \\
8 & 0.218 & 0.516 & 0.220 & 0.474 & 0.436 & 0.534 & 0.462 & 0.557 & 0.475 & 0.563 \\
16 & 0.221 & 0.517 & 0.217 & 0.469 & 0.437 & 0.532 & 0.462 & 0.556 & 0.477 & 0.563 \\
\hline
\end{tabular}
}
\caption{The MRR and Hit@10 results of EANS with small negative samples setting on WN18RR}
\label{tab:small_ns_wn18rr}
\end{table*}

We report MRR and Hits@10 results of 4 different negative sample sizes $n \in \{1,4,8,16\}$ to various knowledge graph embedding models with EANS. Table \ref{tab:small_ns_fb15k237} and Table \ref{tab:small_ns_wn18rr} show the results on FB15K-237 and WN18RR, respectively. Through both results, we confirm that EANS works well even when the negative sample size is extremely reduced.

\end{document}